\documentclass[sigconf]{acmart}

\usepackage{multirow}  
\graphicspath{{figures/}}  

\AtBeginDocument{%
  \providecommand\BibTeX{{%
    \normalfont B\kern-0.5em{\scshape i\kern-0.25em b}\kern-0.8em\TeX}}}

\setcopyright{acmcopyright}
\copyrightyear{2022}
\acmYear{2022}

\acmConference[ACM Conference’22]{Proceedings of ACM International Conference (ACM Conference’22)}{January, 2022}{City, Country}
\acmBooktitle{Proceedings of ACM International Conference’22, January, 2022, City, Country}
\acmPrice{15.00}

\begin{document}

\title{FCL-GAN: A Lightweight and Real-Time Baseline for Unsupervised Blind Image Deblurring}


\author{Suiyi Zhao, Zhao Zhang, Richang Hong}
\affiliation{%
	  \institution{Hefei University of Technology, China}}

\author{Mingliang Xu}
\affiliation{%
	\institution{Zhengzhou University, China}}

\author{Yi Yang}
\affiliation{%
	\institution{University of Technology Sydney, Australia}}

\author{Meng Wang}
\affiliation{%
	\institution{Hefei University of Technology, China}}


\begin{abstract}
  Blind image deblurring (BID) remains a challenging and significant task. Benefiting from the strong fitting ability of deep learning, paired data-driven supervised BID method has obtained great progress. However, paired data are usually synthesized by hand, and the realistic blurs are more complex than synthetic ones, which makes the supervised methods inept at modeling realistic blurs and hinders their real-world applications. As such, unsupervised deep BID method without paired data offers certain advantages, but current methods still suffer from some drawbacks, e.g., bulky model size, long inference time, and strict image resolution and domain requirements. In this paper, we propose a lightweight and real-time unsupervised BID baseline, termed Frequency-domain Contrastive Loss Constrained Lightweight CycleGAN (shortly, FCL-GAN), with attractive properties, i.e., no image domain limitation, no image resolution limitation, 25x lighter than SOTA, and 5x faster than SOTA. To guarantee the lightweight property and performance superiority, two new collaboration units called \textit{lightweight domain conversion unit} (LDCU) and \textit{parameter-free frequency-domain contrastive unit} (PFCU) are designed. LDCU mainly implements inter-domain conversion in lightweight manner. PFCU further explores the similarity measure, external difference and internal connection between the blurred domain and sharp domain images in frequency domain, without involving extra parameters. Extensive experiments on several image datasets demonstrate the effectiveness of our FCL-GAN in terms of performance, model size and inference time. 
\end{abstract}



\maketitle

\quad
	
\section{Introduction}

\begin{figure}[t]
	\centering
	\includegraphics[width=1\columnwidth]{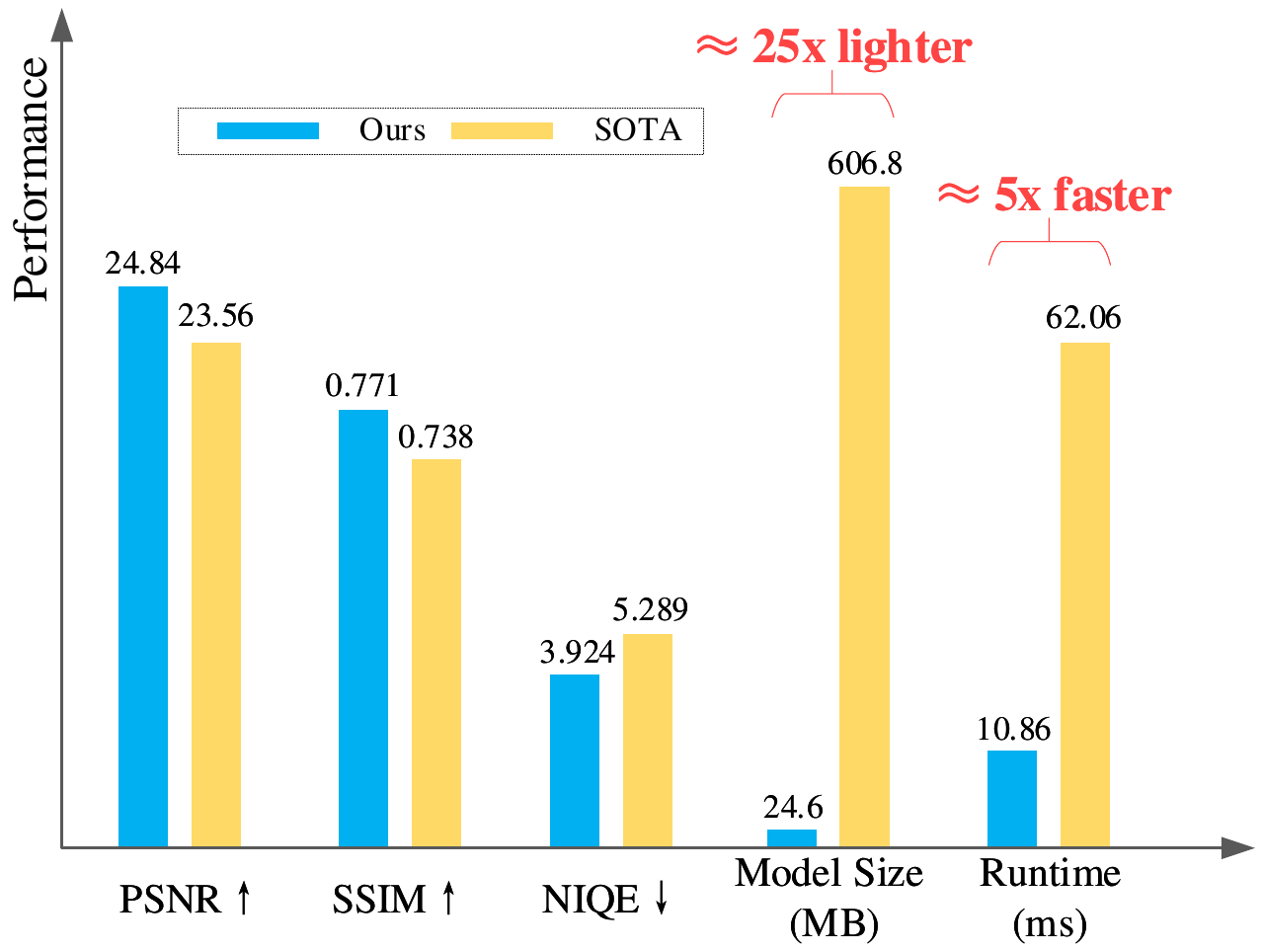}
	\vspace{-6mm}
	\caption{Performance comparison between the SOTA unsupervised method UID-GAN \cite{lu2019unsupervised} and our FCL-GAN. Where \textit{PSNR}, \textit{SSIM} \cite{wang2004image} and \textit{NIQE} \cite{mittal2012no} are compared on the GoPro test set \cite{nah2017deep}, and \textit{"Runtime"} indicates the time to infer a 1280*720 resolution image using Nvidia RTX 2080Ti.}
	\vspace{-4mm}
	\label{fig1}
\end{figure}

\begin{figure*}[t]
	\centering
	\includegraphics[width=2.1\columnwidth]{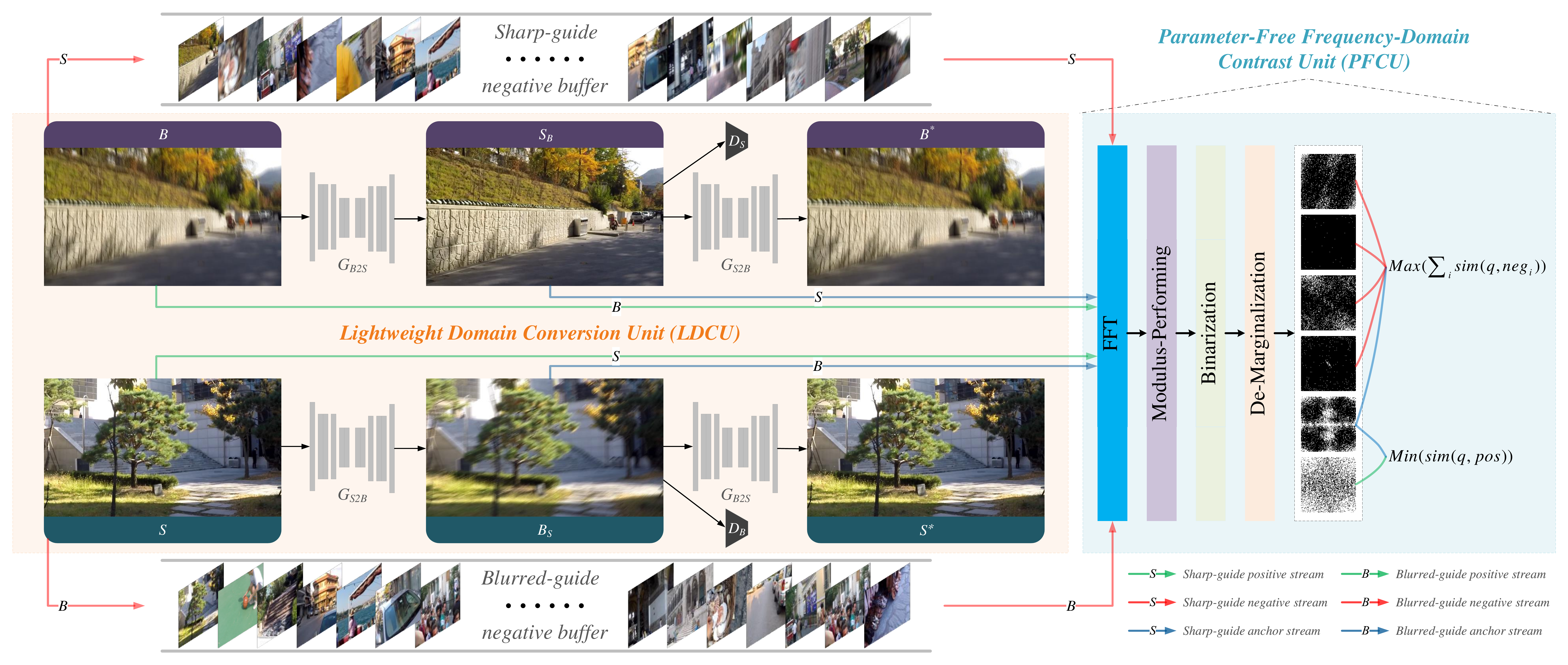}
	\vspace{-4mm}
	\caption{The architecture of FCL-GAN, which consists of two collaborative units: LDCU and PFCU. LDCU performs conversion of different domains (i.e., $B$$\rightarrow$$S_B$$\rightarrow$$B^\ast$$\approx$$B$, $S$$\rightarrow$$B_S$$\rightarrow$$S^\ast$$\approx$$S$), and PFCU pulls similar latent representation (i.e., $blurred$$\rightarrow$$\leftarrow$$blurred$, $sharp$$\rightarrow$$\leftarrow$$sharp$) and pushes dissimilar latent representation (i.e., $blurred$$\leftarrow$$\rightarrow$$sharp$) in the frequency domain.}
	\vspace{3mm}
	\label{fig2}
\end{figure*}

Blind image deblurring (BID), as a classical multimedia processing task, aims at recovering a latent image from a blurred input. Blurred pictures in real-world are common, which will greatly affect the image quality and degrade the related low-level vision perception and high-level tasks. Conventional optimization-based methods assume the latent sharp image satisfies various priors \cite{bai2019single,joshi2009image, pan2014deblurringtext,chen2019blind, pan2016blind,yan2017image}, and transform the deblur problem into maximum a posteriori probability optimization. However, these methods require complex iterative optimizations with a long inference time. Moreover, the deblurring results contain unpleasant heavy artifacts.

In recent years, data-driven deep BID methods (usually supervised) \cite{nah2017deep,tao2018scale,kupyn2018deblurgan,zhang2019deep,suin2020spatially,zhang2020deblurring,chen2021hinet,zamir2021multi,cho2021rethinking,zamir2022restormer} have achieved superior performance, benefiting from the rapid development of deep learning. Data-driven supervised BID methods usually use a large amount of synthetic paired data to train a deep neural network (DNN) with various cost functions, and then learn an end-to-end mapping from blurred to sharp images. Note that a large amount of synthetic paired data is the key to the success of supervised BID methods. However, collecting paired data by hand is expensive, and strictly paired data is usually impossible in reality. Besides, realistic blur is more complex and diverse than the synthetic ones. Moreover, the synthetic data cannot reflect all blurs, resulting in the limitated performance of current supervised deblur methods in practical applications. These challenges have given rise to the unsupervised deep BID methods \cite{nimisha2018unsupervised,lu2019unsupervised,chen2018reblur2deblur,zhao2021unsupervised}, which aim at learning the mapping of blurred images to sharp images without any paired data. 

Deep unsupervised BID methods are rarely studied due to higher difficulty and greater challenge, compared with the supervised models. Specifically, due to the lack of strong constraints in unpaired case, researchers tend to design bulky deep models to seek weak connections between input and output, while this usually results in long inference time. As a result, deploying such bulky deep models into mobile devices for online and real-time computation will be challenging. Besides, current unsupervised deep models perform BID under numerous restrictions, e.g., specific image domain \cite{nimisha2018unsupervised,lu2019unsupervised}, needing multiple frames for training \cite{chen2018reblur2deblur} and small input resolutions \cite{zhao2021unsupervised}. Clearly, these will directly limit the development and real-world application of unsupervised BID models. 

In this paper, we therefore propose a simple and effective deep unsupervised  BID baseline to address the aforementioned shortcomings of current studies. \textit{From the performance point of view, the proposed baseline should satisfy the following conditions}: 1) \textbf{high model performance}, i.e., achieving SOTA performance of unsupervised methods; 2) \textbf{lightweight model size}, i.e., meeting the configuration requirements of most devices at present; 3) \textbf{fast inference speed}, i.e., the baseline can work in a real-time situation, below 33.33 ms/frame. Besides, \textit{from the perspective of overcoming limitations, the baseline should also satisfy the following conditions}:  1) \textbf{no image domain limitation}, i.e., can work on arbitrary image domain; 2) \textbf{no image resolution limitation}, i.e., can process high-resolution images; 3) \textbf{no multi-frame input limitation}, i.e., can perform single-input-single-output mapping. To illustrate our baseline meets the above conditions, we accept a \textit{high-resolution} \textit{natural} image as input and compare the  SOTA's output in Figure \ref{fig1}. Clearly, our method outperforms SOTA in different aspects. 

The major contributions of this paper are summarized as follows:
\begin{itemize}
	\item We propose a lightweight and real-time baseline (FCL-GAN) for unsupervised BID. FCL-GAN drives the task by a frequency-domain contrastive loss constrained lightweight CycleGAN, as shown in Figure \ref{fig2}. Two new collaborative units, called lightweight domain conversion unit (LDCU) and parameter-free frequency-domain contrast unit (PFCU), are designed, which jointly makes sure ``lightweight'' and ``real-time'', in addition to overcoming the limitations of current unsupervised methods. To the best of our knowledge, this is the first lightweight and real-time deep unsupervised BID method, which can be regarded as a guiding method for future research. 
	
	\item Although CycleGAN \cite{zhu2017unpaired} is a popular architecture trained without paired data by a cycle consistency loss, it still suffers from large chromatic aberration, severe artifacts and model redundancy when handling BID task (See Figure \ref{fig3}). As such, we disassemble and analyze the model structure and basic constituent units, and present a novel lightweight domain conversion unit (LDCU), which can deblur the degraded images better in a lightweight and real-time manner.
	
	\item To further improve the deblur performance without extra parameters, we introduce a new concept of frequency-domain contrastive learning (FCL), and present a parameter-free frequency-domain contrastive unit (PFCU) to measure the similarity between latent representations in frequency domain as a contrastive constraint. FCL can also address the shortcomings of contrastive learning on deblur task.
	
	\item Extensive simulations on several datasets demonstrated the SOTA unsupervised deblur performance of our FCL-GAN, with stonger generalization ability to handle real-world blurs. To be specific, FCL-GAN allows for a smaller model size (24.6MB vs. 606.8MB) and less inference time  (0.011s vs. 0.062s), compared with current SOTA method.
\end{itemize}

\section{Related Work}

\subsection{Data-Driven Deep BID}

\textbf{Deep fully-supervised BID.}
Benefiting from large-scale paired training data, supervised BID methods can learn accurate mapping more easily. For example, a ``multi-scale'' end-to-end training manner \cite{nah2017deep} was proposed for directly deblurring without blur kernels. Kupyn et al. \cite{kupyn2018deblurgan, kupyn2019deblurgan} leverages CGAN \cite{mirza2014conditional} and wgan-gp \cite{gulrajani2017improved} to obtain visually realistic deblurring results in a generative-adversarial manner. Zhang et al. \cite{zhang2020deblurring} provide a Real-World blurred images dataset and design an effective GAN-based network (DBGAN) to model real-world blur. The ``multi-patch'' strategy is also used to partition the image in spatial dimensions and then perform a coarse-to-fine progressive deblurring \cite{zhang2019deep}\cite{zamir2021multi}. Instead of using convolution, Zamir et al. \cite{zamir2022restormer} introduce and refine the transformer \cite{vaswani2017attention} to deblur high-resolution images, achieving state-of-the-art (SOTA) fully-supervised BID performance.

\textbf{Deep semi-supervised BID.}
Semi-supervised methods learn an approximate blurred-sharp mapping based on small-scale paired data and large-scale unpaired data. Compared with fully-supervised BID, it will be more difficult for the semi-supervised BID to learn a blurred-sharp mapping. Therefore, Nimisha et al. \cite{nimisha2018semi} first try to estimate the global camera motion from small-scale paired data in a semi-supervised manner and use the obtained global camera motion to perform single image deblurring and change detection.

\textbf{Deep unsupervised BID.}
The training of unsupervised methods does not involve paired data, and instead, it uses large-scale unpaired data for deblurring. Compared with the fully-/semi-supervised BID, unsupervised BID methods are more difficult to learn an accurate blurred-sharp mapping due to the weak constraint between blurred and sharp images. For example, a self-supervised optimization scheme \cite{chen2018reblur2deblur} was proposed based on the existing deblur models, which utilizes the continuous frames of video and introduces a physically-based blur information model during training to improve the deblur performance. Strictly speaking, this is not an unsupervised BID technique since it is built on a fully supervised model and continuous frames. Based on the simple generative adversarial network (GAN) \cite{goodfellow2014generative}, Nimisha et al. \cite{nimisha2018unsupervised} introduce scale-space gradient loss and reblurring loss to self-supervise the model to perform domain-specific deblurring. Lu et al. \cite{lu2019unsupervised} entangle the content and blur of the blurred image over the domain-specific datasets, and then the blur can be easily removed from the blurred image. Zhao et al. \cite{zhao2021unsupervised} focus on the chromatic aberration problem of unsupervised methods and proposes blur offset estimation and adaptive blur correction strategy to maintain color information while deblurring, which achieves better unsupervised BID performance.

\subsection{Contrastive Learning}
With the iteration of self-supervised and unsupervised techniques, contrastive learning has been attracting more and more attention. In general, contrastive learning aims to map the original data into a latent representation space in which anchors are pulled close to positive samples and pushed farther away from negative samples. In this way, the model can not only learn from positive signals but also benefits from correcting undesirable behaviors. In recent years, Contrastive learning has been widely used in various high-level vision tasks, e.g., object detection \cite{xie2021detco}, medical image segmentation \cite{chaitanya2020contrastive} and image caption \cite{dai2017contrastive}, which has achieved superior performance in high-level tasks and is receiving increasing attention. More recently, Contrastive learning has been successfully applied to various low-level vision tasks and achieved SOTA performance, e.g., image denoising \cite{dong2021residual}, image dehazing \cite{wu2021contrastive} and image super-resolution \cite{zhang2021blind}. Similarly, Chen et al. \cite{chen2022unpaired} first introduce contrastive learning to unsupervised single image deraining task and perform contrastive constraints at the feature level, which achieves SOTA performance in unsupervised single image deraining.

\section{Proposed Baseline Method}
\subsection{Architecture}
We show the architecture and learning process of our FCL-GAN in Figure \ref{fig2}. Clearly, it has two main cooperating units, i.e., LDCU and PFCU. LDCU is a lightweight unit that implements interconversion between different domain images. PFCU applies contrastive learning for deblurring at a technical level, making the output anchor closer to the positive sample. In what follows, we detail the interactive process between the LDCU and the PFCU. For ease of description, we will begin with some basic definitions: 

\begin{itemize}
	\item \textbf{``Stream''}: The process of data transfer, e.g., ``positive stream'' is used to transfer positive samples and positive latent representations. When the positive samples are sharp, the ``stream'' will be denoted as ``sharp-guide stream''.
	
	\item \textbf{``Buffer''}: The place where the samples used by the previous epochs are stored. Samples inside the ``buffer'' are treated as negative samples. For example, for the ``sharp-guide'' streams, samples inside the ``buffer'' are all blurred. 
\end{itemize}

LDCU is the basis of the whole architecture of FCL-GAN, which implements domain conversion. As shown in Figure \ref{fig2}, the LDCU contains two branches, i.e., the deblur branch: $B$$\rightarrow$$S_B$$\rightarrow$$B^\ast$$\approx$$B$ and the reblur branch:  $S$$\rightarrow$$B_S$$\rightarrow$$S^\ast$$\approx$$S$. Similarly, PFCU will impose different contrastive constraints for the two branches, i.e., sharp-guide contrast constraint for the deblur branch, while using blurred-guide contrast constraint for the reblur branch. Let's take the deblur branch as an example, the latent images $S_B$, the sharp images $S$ and the images in the sharp-guide negative buffer will be transferred to the PFCU as the anchor, positive samples and negative samples, respectively, to obtain the latent representations. Then, the similarity are calculated and contrastive constraints are performed. Note that the framework of FCL-GAN is lightweight and real-time, since LDCU is lightweight and real-time, and PFCU does not involve extra parameters and calculations during the inference process.

\subsection{Lightweight Domain Conversion Unit}
Let $\mathbb{D}_B$ be the blurred image domain, and let $\mathbb{D}_S$ be the sharp image domain, our ultimate goal is to map the images in $\mathbb{D}_B$ without ground truth to $\mathbb{D}_S$. To this end, we introduce LDCU as the backbone of our FCL-GAN. LDCU contains two generators ($G_{B2S}$ and $G_{S2B}$) and two discriminators ($D_B$ and $D_S$): $G_{B2S}$ ($G_{S2B}$) for mapping images in $\mathbb{D}_B$ ($\mathbb{D}_S$) to $\mathbb{D}_S$ ($\mathbb{D}_B$); $D_B$ ($D_S$) for discriminating the authenticity of images in $\mathbb{D}_B$ ($\mathbb{D}_S$). Motivated by \cite{wei2021deraincyclegan, chen2022unpaired}, two functional circuits are set to perform deblurring and blur generation. Taking the deblurring circuit as an example, given images $B$ in $\mathbb{D}_B$, $G_{B2S}$ maps $B$ to images $S_B$ in $\mathbb{D}_S$, and then $G_{S2B}$ remaps $S_B$ to images $B^\ast$ in $\mathbb{D}_B$, i.e., $B$$\rightarrow$$S_B$$\rightarrow$$B^\ast$$\approx$$B$. Throughout the deblurring circuit, $D_S$ is used to discriminate whether $S_B$ is truly an image in $\mathbb{D}_S$. Similar to the deblurring circuit, the blur generation circuit accepts images in $\mathbb{D}_S$ for the opposite mapping, i.e., $S$$\rightarrow$$B_S$$\rightarrow$$S^\ast$$\approx$$S$.

\begin{figure}[t]
	\centering
	\includegraphics[width=0.9\columnwidth]{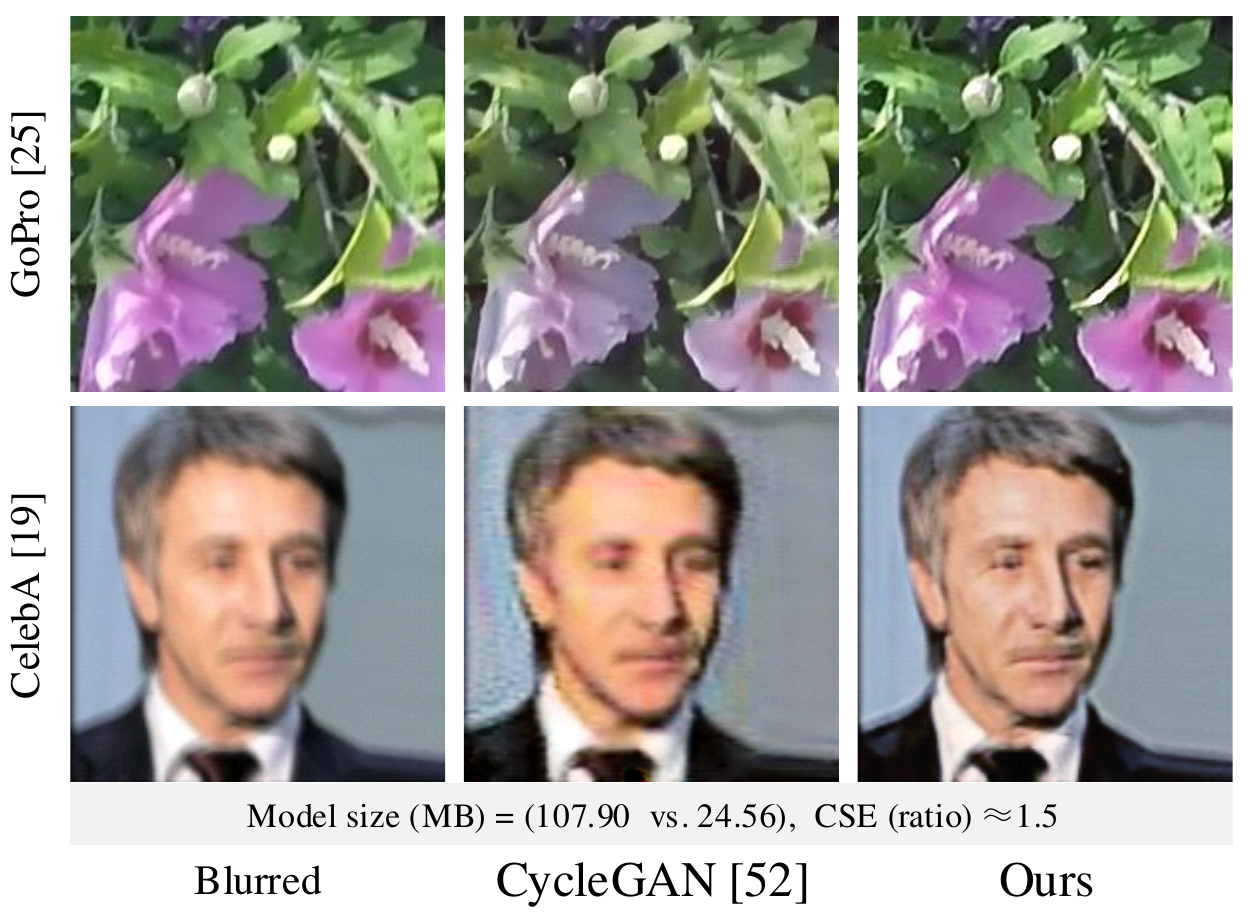}
	\vspace{-2mm}
	\caption{Visualizing the negative effects when CycleGAN \cite{zhu2017unpaired} is applied directly to the deblurring task. We can see that CycleGAN often results in severe chromatic aberrations (both two groups are about 1.5 times larger than ours), model redundancy (107.90 MB vs. Our 24.56 MB) and artifacts.}
	\vspace{-4mm}
	\label{fig3}
\end{figure}

In general, converting images between $\mathbb{D}_B$ and $\mathbb{D}_S$ can be regarded as an image-to-image translation task. As a classical image-to-image translation framework in this domain, CycleGAN \cite{zhu2017unpaired} has a very powerful ability to learn inter-domain differences. However, CycleGAN cannot achieve the expected deblurring result, and there may be several potential reasons for this: 1) \textbf{Incompatibility of network architectures}. Unlike the image-to-image translation task, the network architecture of the deblurring method is elaborated and sometimes a slight change can severely disrupt the results, e.g., changing BN to IN, which can be seen in ablation studies; 2) \textbf{Different degrees of task complexity}. Image to image translation tasks (e.g., zebra$\leftrightarrow$horse, orange$\leftrightarrow$apple and summer$\leftrightarrow$winter) tend to have a straightforward inter-domain difference, so the deep network can easily learn precise inter-domain difference. However, the inter-domain difference for image deblurring is usually complex, which can be reflected by the blurring degree. As exemplified by the blurring degree in extreme cases, when the blurring degree is very large, one cannot even read any information in the blurred image; when the blurring degree is very small, the blurred image can even be regarded as a sharp image. Therefore, directly migrating CycleGAN to the deblurring task is not feasible and will lead to a series of negative effects, such as chromatic aberration, severe artifacts and parameter redundancy, as shown in Figure \ref{fig3}. 

As such, we strive to make the network lightweight and more effective. Specifically, we perform a fully recursive decomposition of entire LDCU, carefully design each minimal component, and finally reconstitute LDCU with these elaborate components. We will mainly focus on refining the generator in LDCU, since it takes up the direct representation of domain conversion, as the key component of LDCU. Generally speaking, the number of generator parameters is much higher than that of the discriminator in the whole generation-discrimination structure because a sufficient number of parameters will better exploit the powerful nonlinear mapping capability of DNN. However, the generator will tend to saturate when the number of parameters reaches a certain level. Thus, a larger number of parameters will make the model redundant, which will seriously hinder the inference and deployment of the model.

To minimize model redundancy, we introduce and perform the following operations from bottom to top: 1) \textbf{meta design;} 2) \textbf{lightweight structure design.} Next, we detail the operations. 

\textbf{Meta design.} Convolutional neural networks (CNN) consist of a stack of non-divisible units, e.g., convolution (Conv), batch normalization (BN), instance normalization (IN) and Rectified linear unit (ReLU). However, efficiently organizing these non-divisible units to build a deblurring network is a problem worth exploring. Conv and ReLU are necessary, because they respectively support the CNN's parametric learning capability and nonlinear fitting capability. Note that the normalization unit is a mandatory factor.

\begin{figure*}[t]
	\centering
	\includegraphics[width=1.95\columnwidth]{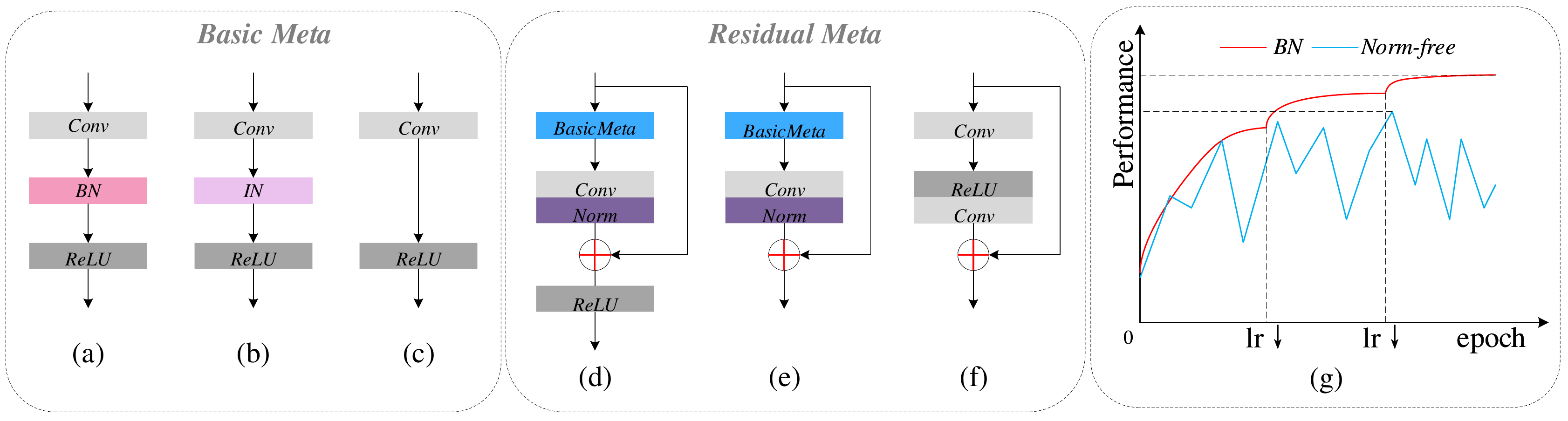}
	\vspace{-2mm}
	\caption{Different forms of basic and residual metas. "Norm" means the normalization in basic beta, which can be BN, IN or norm-free. LDCU is based on (a) and (e). (g) is a schematic diagram for norm-free that leads to instability during training.}
	\vspace{-2mm}
	\label{fig4}
\end{figure*}

For supervised deblurring, researchers usually did not use the normalization units for model's structure design, since adding additional normalization units under the condition of adequate constraints may severely hinder the model to learn an accurate \textit{1-to-1:} blurred$\rightarrow$sharp mapping \cite{nah2017deep,tao2018scale,zhang2019deep,park2020multi,tsai2021banet,zamir2021multi,cho2021rethinking}. However, in some other fields (e.g., image-to-image translation and style transfer), researchers also used IN for model's structure design, because these tasks expect to learn a \textit{1-to-1} style mapping \cite{zhu2017unpaired,yi2017dualgan,zhang2019disentangled}. However, in unsupervised deblurring, the lack of strong constraints and the diversity of blurring degrees lead to \textit{n-to-n} style mapping. Besides, BN normalizes to multiple samples (mini-batch) instead of IN, and is more likely to learn global differences among domains. To this end, we introduce the concepts of basic and residual metas to learn this weakly-constrained style mapping based on BN, as shown in Figure \ref{fig4} (a) and (e). We have analyzed the importance of our proposed basic and residual metas in detail in ablation studies.

\begin{figure}[t]
	\centering
	\includegraphics[width=0.85\columnwidth]{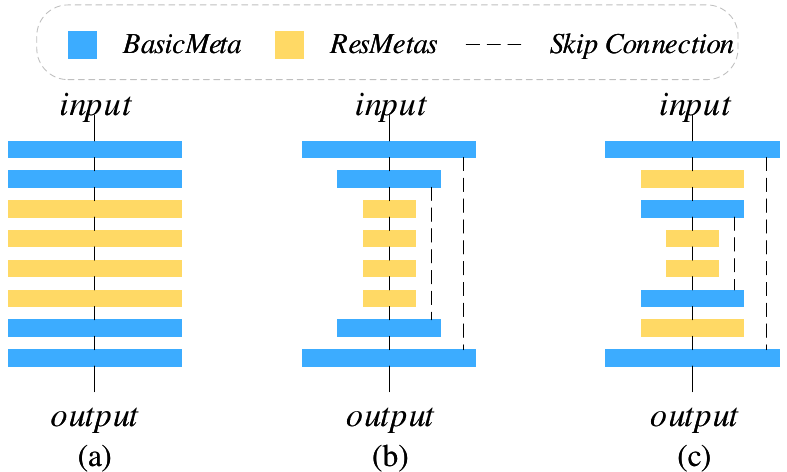}
	\vspace{-2mm}
	\caption{Different structures of DNN models. ResMetas contains several residual metas. Only two basic metas are shown when coding for simplifying the schematic.}
	\vspace{-4mm}
	\label{fig5}
\end{figure}

\textbf{Lightweight structure design.} In the field of image restoration, many efforts have been devoted to designing complex models to obtain desired results. However, no matter how complex the model is, they are all based on two underlying structures, i.e., \textit{encode-decoder structure} \cite{nah2017deep} (see Figure \ref{fig5} (b)) or \textit{single-scale pipeline}  \cite{ren2019progressive} (see Figure \ref{fig5} (a)). The encoder-decoder structure can effectively abstract the content information but cannot maintain the spatial details of the images. The single-scale pipeline can ensure accurate spatial information, but cannot easily abstract the image contents. With the same settings, the number of parameters is the same as that of encoder-decoder structure. Nevertheless, from the viewpoint of inference speed, without the encoding-decoding process, the model based on single-scale pipeline forward at the original resolution, which can reduce the inference speed and affect the real-time capability. To simplify the descriptions, we next use \textit{encode-decoder} and \textit{single-scale} to represent these two structures. 

Considering the strengths and weaknesses of the two existing structures, Zamir et al. \cite{zamir2021multi} add an additional original resolution module after encoder-decoder, which balances the advantages of both, but also involves extra parameters and inference time. Inspired by Cho et al. \cite{cho2021rethinking}, we introduce a new structure termed \textit{lightweight encoder-decoder (LED)} to incorporate the strengths of both encoder-decoder and single-scale, as shown in Figure \ref{fig5} (c). Compared with \textit{encoder-decoder}, \textit{LED} stacks residual metas at a larger resolution, which will be more favorable for preserving the spatial details. Compared with \textit{single-scale}, \textit{LED} contains both encoding and decoding, which facilitates the abstraction of the image content.

\begin{figure}[t]
	\centering
	\includegraphics[width=1\columnwidth]{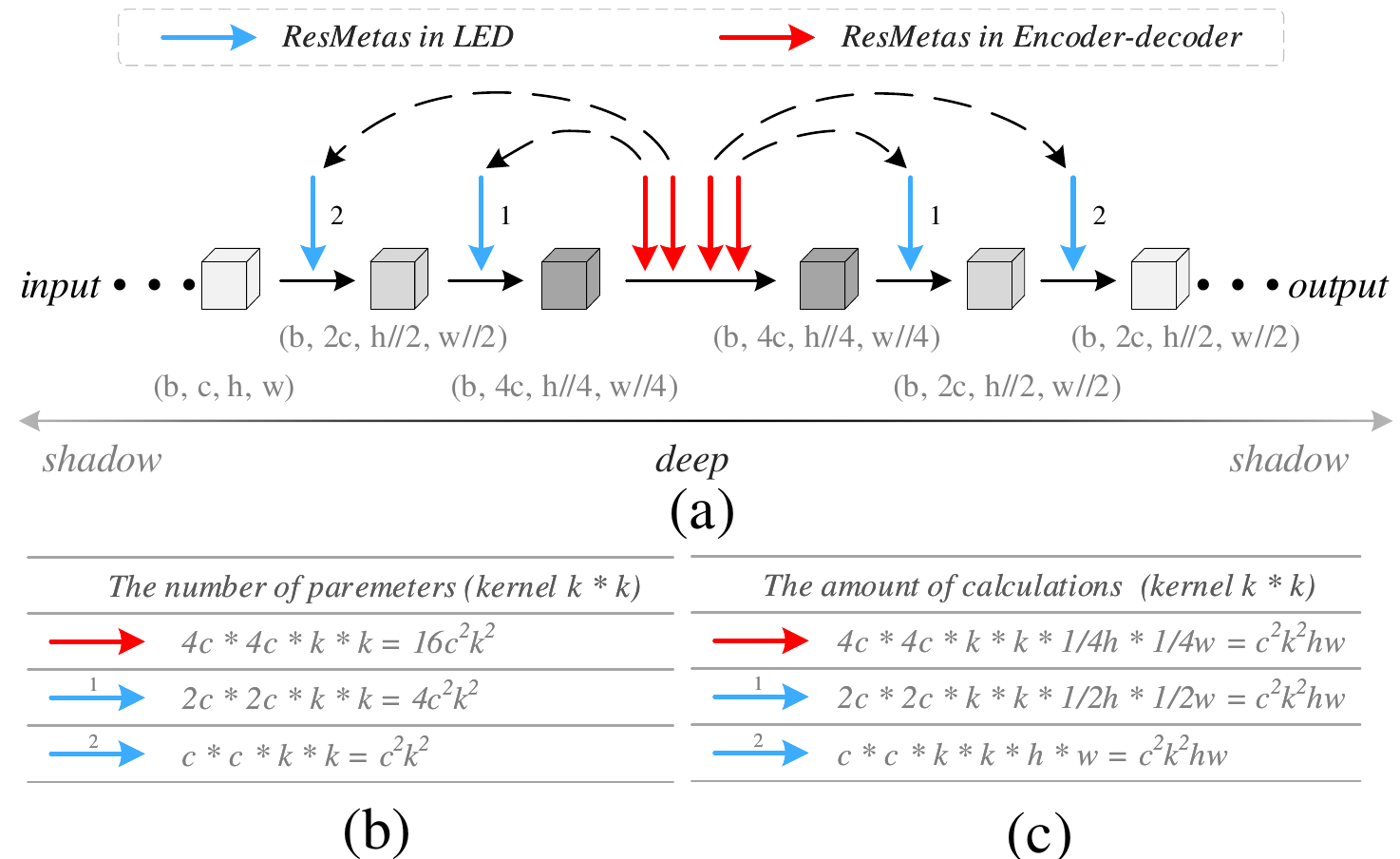}
	\vspace{-5mm}
	\caption{(a) Difference between \textit{LDE} and \textit{encoder-decoder}. (b) and (c) compare the number of parameters and the number of calculations brought by the red and blue arrows.}
	\vspace{-5mm}
	\label{fig6}
\end{figure}

Figure \ref{fig6} (a) illustrates the difference between \textit{LED} and \textit{encoder-decoder}, i.e., the ResMetas of \textit{encoder-decoder} are all gathered in the deep layer, while our \textit{LED}'s can be distributed in the shallow layer. Figures \ref{fig6} (b) and (c) compare the number of parameters and the number of calculations for both structures under the same setting. Note that it is known that for the number of parameters: \textit{encoder-decoder}'s = \textit{single-scale}'s; for the number of calculations: \textit{encoder-decoder}'s < \textit{single-scale}'s. However, as we can see from Figure \ref{fig6}, LED's parameters are much lower than that of the encoder-decoder and the number of calculations is the same. At this point, we can conclude that \textit{LED} considers the advantages of \textit{encoder-decoder} and \textit{single-pipeline} while making the model lightweight without introducing extra calculations. Finally, we will illustrate the effectiveness of our \textit{LED} in ablation studies.

\subsection{Parameter-Free Frequency-Domain Contrastive Unit (PFCU)}
The difference between blurred and sharp images is hard to explain in the spatial domain. However, the difference in the frequency domain can be explained by the fact that the blurred image loses the high-frequency signal in a sharp image. A few efforts have been made to investigate the frequency domain in the BID task, but they are limited to only two ways: either using constraints in the frequency domain between the latent image and ground truth \cite{cho2021rethinking}, or integrating Fast Fourier transform (FFT) and inverse Fast Fourier transform (IFFT) in the model \cite{mao2021deep}. However, these approaches are ground truth-dependent and difficult to understand.

We have studied a large number of blurred and sharp images in the frequency domain and found that: without normalizing the images, the blurred images tended to be all black, however the sharp images tended to be all white. Besides, the higher the degree of blurring, the darker the image, as shown in Figure \ref{fig7}. However, it is very challenging to directly measure the difference between blurred and sharp frequency domain images in an unsupervised manner. Therefore, we put our hopes on powerful contrastive learning.

Two key questions need to be addressed to fully exploit the effect of contrastive learning: 1) How to get the latent representation of samples? 2) How to calculate the similarity between the latent representations? Note that DCD-GAN \cite{chen2022unpaired} used additional branch to obtain potential representation and used the cosine similarity function to define distances. However, this approach will make the model tend to update the extra branches and ignore the task itself.

To address the above problems well, based on the principle of lightweight, we introduce a novel parameter-free frequency-domain contrast unit (i.e., PFCU) for contrastive learning in frequency domain. Figure \ref{fig2} shows the structure of PFCU in detail, which consists of several layers, i.e., the FFT layer, modulus-performing layer, binarization layer and de-marginalization layer. Specifically, given an input sample $I$, the FFT layer firstly performs a fast Fourier transform on $I$ to obtain the frequency domain output $I_f$ of complex type. The modulus-performing layer then modulos $I_f$ to obtain the real space representation $I_r$. The binarization layer aims to binarize $I_r$ according to a threshold value (zero) to obtain binarized $I_b$. Finally, the de-marginalization layer aims to preserve $I_b$'s central region to further highlight the frequency domain variation and obtain the latent representation $P_I$ of input $I$. We express the whole process of obtaining the latent representation by the following formula: 

\begin{figure}[t]
	\centering
	\includegraphics[width=1\columnwidth]{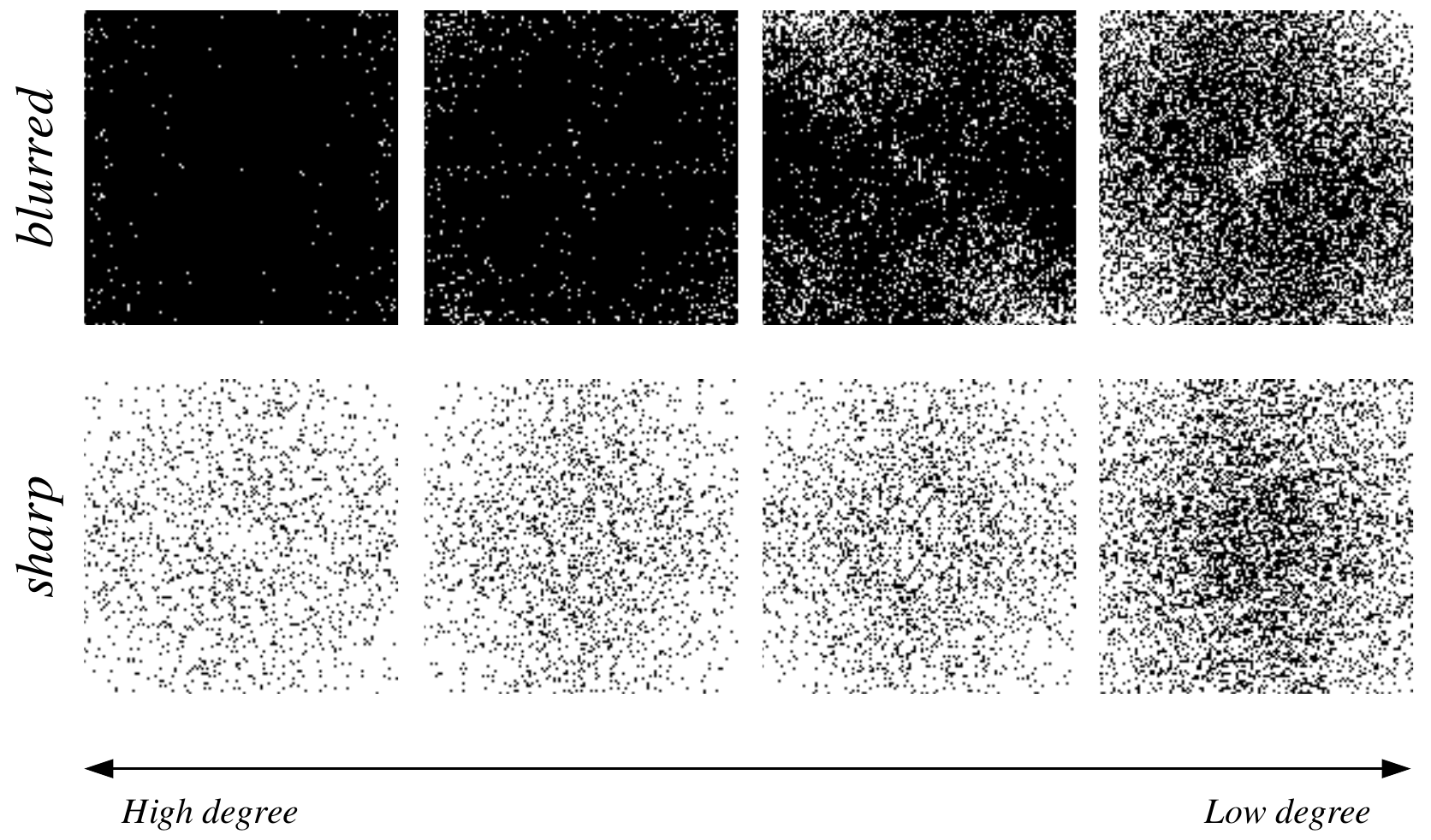}
	\vspace{-5mm}
	\caption{Different degrees of blurred and sharp images in the frequency domain without normalization.}
	\vspace{-4mm}
	\label{fig7}
\end{figure}

\begin{equation}\label{eqn1}
	P_I = PFCU(I).
\end{equation}

\begin{figure}[t]
	\centering
	\includegraphics[width=1\columnwidth]{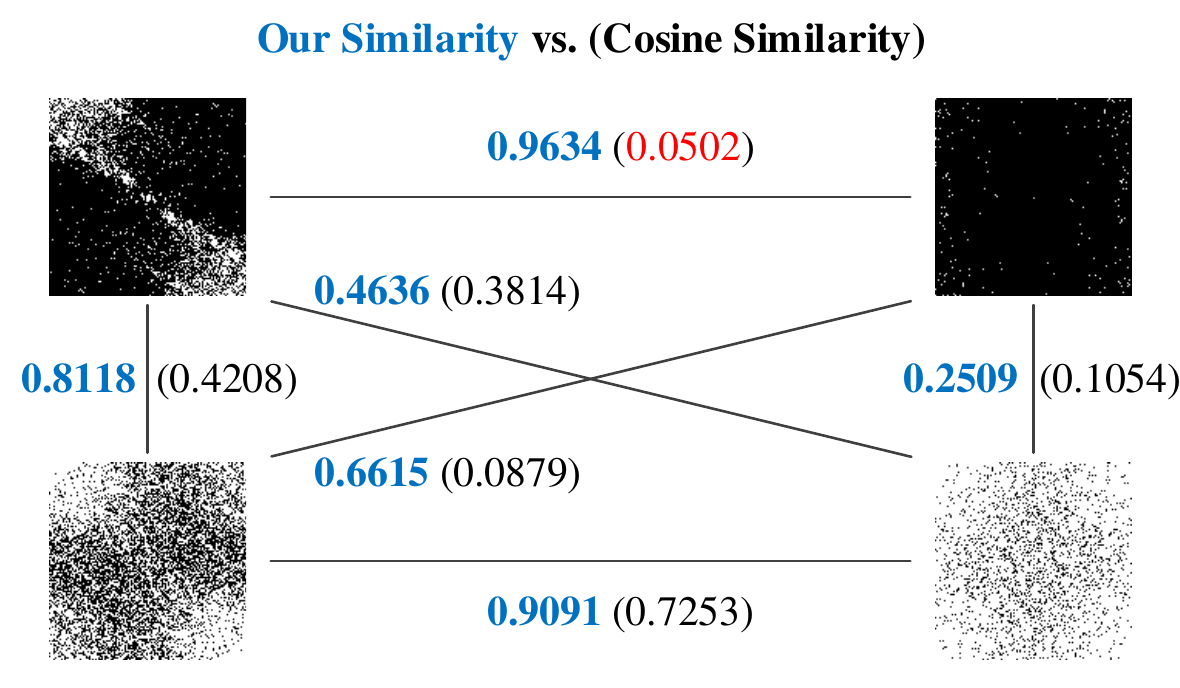}
	\vspace{-6mm}
	\caption{Validation of our similarity against cosine similarity in the frequency domain. The numbers on lines is the similarity between the two representations. The most unreasonable value of the cosine similarity is highlighted in red.}
	\vspace{-2mm}
	\label{fig8}
\end{figure}

After obtaining the potential representation of samples, calculating the similarity will be the most critical issue to be addressed. Note that it is unfeasible to use the traditional cosine similarity directly because there are no learnable parameters in the calculation of latent representations, and two frequency domain images with small differences may have a low similarity score, which is certainly unreasonable (See Figure \ref{fig8}). For this purpose, considering the difference in black coverage between various latent representations, we design a new similarity measurement method. Specifically, given two latent representations $P_1$, $P_2$ $\in$ $\mathbb{R}^{h\times w}$ and a hyperparameter $\omega$, we first chunk $P_1$, $P_2$ to obtain $P_{b1}$, $P_{b2}$ $\in$ $\mathbb{R}^{\frac{h\times w}{\omega^2}\times \omega \times \omega}$, and then calculate the black coverage for each chunk to obtain $R_{b1}$, $R_{b2}$ $\in$ $\mathbb{R}^{\frac{h\times w}{\omega^2}}$. The black coverage for each chunk is calculated as follows:

\begin{equation}\label{eqn2}
	R_b = \frac{\sum_{i=0}^{\omega}\sum_{j=0}^{\omega}(1-b_{i,j})}{\omega^2},
\end{equation}
where $b$$\in$$\mathbb{R}^{\omega\times \omega}$ is one chunk of the chunked latent representation (e.g., $P_{b1}$ and $P_{b2}$ above), $b_{i,j}$ is the $i$-th row and $j$-th column element of $b$, and $\omega$ are the size of $b$, respectively.

Then, we can define and calculate the similarity based on the black coverage $R_{b1}$ and $R_{b2}$ in frequency domain as follows:

\begin{equation}\label{eqn3}
	Sim = 1 - MSE(R_{b1}, R_{b2}).
\end{equation}

\begin{table*}[t]
	\centering
	\caption{Performance comparison on two benchmark datasets: GoPro \cite{nah2017deep} and HIDE \cite{shen2019human}.}
	\vspace{-2mm}
	\setlength{\tabcolsep}{0.7mm}{
		\begin{tabular}{l|l|ccc|cc|c|c}
			\hline
			\multicolumn{2}{c|}{Datasets} & \multicolumn{3}{|c|}{GoPro} & \multicolumn{2}{|c|}{HIDE} & \multirow{2}{*}{Runtime (ms)} & \multirow{2}{*}{Model Size (MB)}\\
			\cline{1-7}
			\multicolumn{2}{c|}{Metrics}  & PSNR/SSIM$\uparrow$ & CSE (Ratio)$\downarrow$ & NIQE$\downarrow$ & PSNR/SSIM$\uparrow$ & CSE (Ratio)$\downarrow$ &  &  \\
			\hline
			& Gong et al. \cite{gong2017motion} & 26.40/0.863 & / & / & / & / & / & 39.00 \\
			Deep supervised & DeepDeblur \cite{nah2017deep} & 29.08/0.914 & / & 4.921 & 25.73/0.874 & / & 4330 & 89.40 \\
			methods & SRN \cite{tao2018scale} & 30.26/0.934 & / & 4.834 & 28.36/0.915 & / & 1870 & 27.50 \\
			\hline
			& CycleGAN \cite{zhu2017unpaired} & 22.54/0.720 & 2.676 & 4.359 & 21.81/0.690 & 3.300 & 12.55 & 107.90 \\ 
			Deep unsupervised & DualGAN \cite{yi2017dualgan} & 22.86/0.722 & 4.384 & 4.176 & / & / & / & 324.20 \\
			methods & UID-GAN \cite{lu2019unsupervised} & 23.56/0.738 & 1.532 & 5.289 & 22.70/0.715 & 2.147 & 62.06 & 606.80 \\
			& Ours & \bfseries{24.84/0.771} & \bfseries{1} & \bfseries{3.924} & \bfseries{23.43/0.732} & \bfseries{1} & \bfseries{10.86} & \bfseries{24.56} \\
			\hline
	\end{tabular}}
	\vspace{-1mm}
	\label{table1}
\end{table*}

\begin{figure*}[t]
	\centering
	\includegraphics[width=1.98\columnwidth]{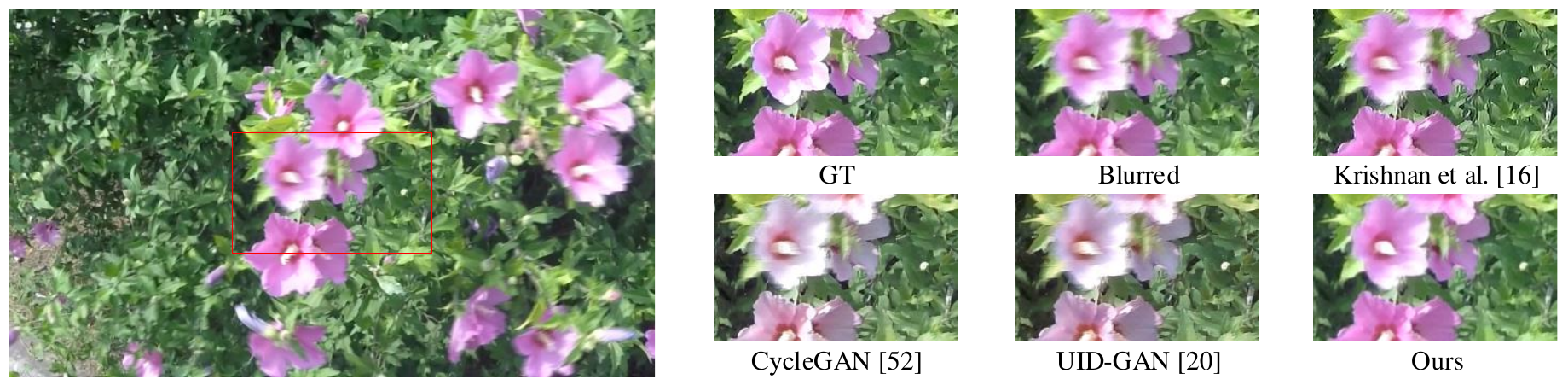}
	\vspace{-2mm}
	\caption{Visual comparison of image deblurring on the GoPro dataset \cite{nah2017deep}.}
	\vspace{-2mm}
	\label{fig9}
\end{figure*}

\begin{figure*}[t]
	\centering
	\includegraphics[width=2\columnwidth]{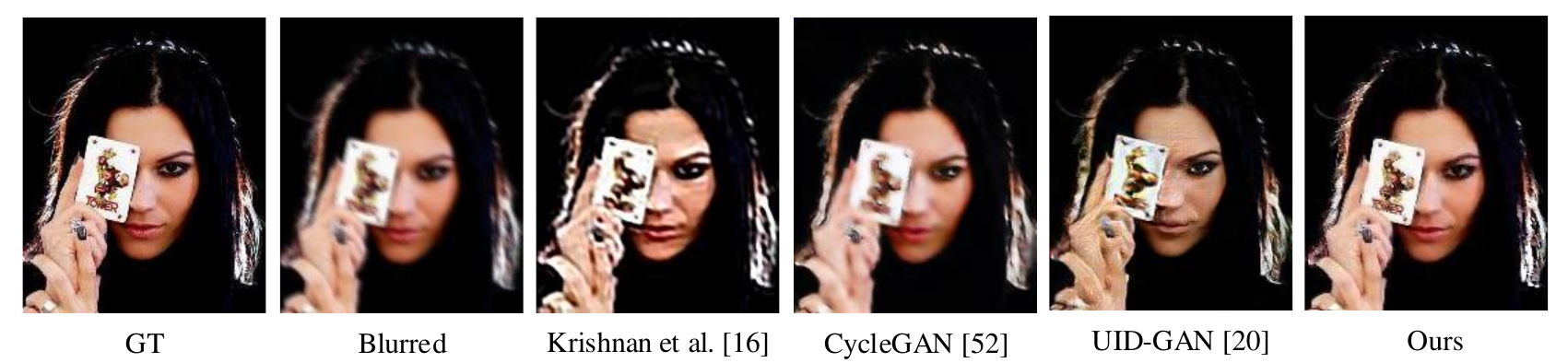}
	\vspace{-3mm}
	\caption{Visual comparison of image deblurring on the CelebA dataset \cite{liu2015deep}.}
	\vspace{-1mm}
	\label{fig10}
\end{figure*}

For ease of understanding, we represent the entire process of calculating the similarity between two different latent representations ($P_1$, $P_2$) by $sim(p_1,p_2)$. We show some results to demonstrate the effectiveness of our similarity measurement method against the coisine similarity in Figure \ref{fig8}. After solving the latent representation acquisition and similarity measure problems, we can easily apply contrastive learning to our FCL-GAN. Specifically, given an anchor $p$, a positive exemplar $p^+$ and several negative exemplars $p^-$, we include the following contrastive loss in training:

\begin{table}[t]
	\centering
	\caption{Performance comparison on CelebA \cite{liu2015deep}.}
	\vspace{-2mm}
	\setlength{\tabcolsep}{0.7mm}{
		\begin{tabular}{l|l|cc}
			\hline
			\multicolumn{2}{c|}{Metrics} & PSNR/SSIM$\uparrow$ & CSE (Ratio)$\downarrow$ \\
			\hline
			& Pan et al. \cite{pan2014deblurringface} & 15.16/0.380 & / \\
			& Xu et al. \cite{xu2013unnatural} & 16.84/0.470 & / \\
			Optimization- & Pan et al. \cite{pan2014deblurringtext} & 17.34/0.520 & / \\
			based methods & Pan et al. \cite{pan2016blind} & 17.59/0.540 & / \\
			& Krishnan et al. \cite{krishnan2011blind} & 18.51/0.560 & / \\
			\hline
			Deep supervi- & DeblurGAN \cite{kupyn2018deblurgan} & 18.86/0.540 & / \\
			sed methods & DeepDeblur \cite{nah2017deep} & 18.26/0.570 & / \\
			\hline
			& CycleGAN \cite{zhu2017unpaired} & 19.40/0.560 & 1.918 \\
			Deep unsuper- & UID-GAN \cite{lu2019unsupervised} & 20.81/0.650 & 2.279 \\
			vised methods & Ours & \bfseries{21.07/0.652} & \bfseries{1} \\
			\hline
	\end{tabular}}
	\label{table2}
\end{table}

\begin{equation}\label{eqn4}
	\mathcal{L}_{ctst}(p,p^+,p^-) = -log[\frac{e^{sim(p,p^+)/\tau}}{e^{sim(p,p^+)/\tau}+\sum_{n=1}^{N}e^{sim(p,p^-)/\tau}})],
\end{equation}
where $N$ is the number of negative exemplars and $\tau$ is the temperature coefficient that is set to 0.07 in all experiments.

\subsection{Loss function}
Except for the contrastive loss $\mathcal{L}_{ctst}$, we also introduce adversarial loss $\mathcal{L}_{adv}$, cycle-consistency loss $\mathcal{L}_{cc}$ and TV regularization $\mathcal{L}_{tv}$ for deblurring. $\mathcal{L}_{tv}$ is applied only to restored sharp images and other losses to both the sharp and blurred domains. The total loss function is: $\mathcal{L}_{total}$$=$$\mathcal{L}_{adv}$$+$$10*\mathcal{L}_{cc}$$+$$0.1*\mathcal{L}_{ctst}$$+$$0.1*\mathcal{L}_{tv}$.

\section{Experiments}
\subsection{Experimental Settings}
\noindent\textbf{Datasets.} In this paper, we evaluate each BID method on four widely-used image datasets, incuding a natural image dataset GoPro \cite{nah2017deep}, a human-aware blurring dataset HIDE \cite{shen2019human}, a human face dataset CelebA (domain-specific) \cite{liu2015deep}, and a real-world blur dataset RealBlur (namely, RealBlur-J and RealBlur-R) \cite{rim2020real}.

\noindent\textbf{Evaluation Metrics.} We use two widely-used reference metrics (PSNR and SSIM \cite{wang2004image}), one non-reference metric (NIQE \cite{mittal2012no}) and one chromatic aberration metric (i.e., color sensitive error, CSE \cite{zhao2021unsupervised}) for evaluations. We also compare the model size to measure the lightweight property. Besides, to measure the real-time performance of each model, we compare the inference time on Nvidia RTX 2080 Ti. In our experimental results: the symbol $\uparrow$ means the higher, the better, while the symbol $\downarrow$ means the lower, the better.

\noindent\textbf{Compared Methods.} We compare our FCL-GAN with 13 methods, including five optimization-based methods: \cite{pan2014deblurringface}, \cite{pan2014deblurringtext}, \cite{pan2016blind}, \cite{xu2013unnatural}, \cite{krishnan2011blind}; four data-driven deep supervised methods:\cite{gong2017motion}, DeblurGAN \cite{kupyn2018deblurgan}, DeepDeblur \cite{nah2017deep}, SRN \cite{tao2018scale}; three data-driven deep unsupervised methods: DualGAN \cite{yi2017dualgan}, CycleGAN \cite{zhu2017unpaired}, UID-GAN \cite{lu2019unsupervised}. We prefer to use the pre-trained model. However, if the settings are inconsistent or there is no pre-trained model, we will retrain it using the code provided by the authors.

\begin{figure*}[t]
	\centering
	\includegraphics[width=2\columnwidth]{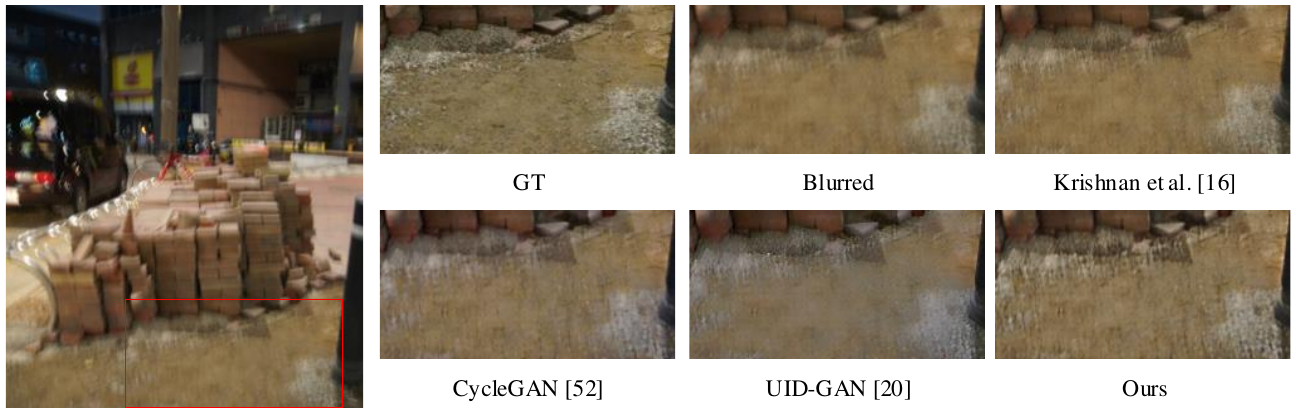}
	\vspace{-1mm}
	\caption{Visual comparison of image deblurring on the RealBlur-J dataset\cite{rim2020real}.}
	\vspace{0mm}
	\label{fig11}
\end{figure*}

\noindent\textbf{Implementation Details.} The proposed baseline model is implementated based on PyTorch version 1.10 and Nvidia RTX 3090 with 24G memory. We set 80 epochs for training, using Adam \cite{kingma2014adam} with $\beta_1$=0.5 and $\beta_2$=0.999 for optimization. The initial learning rate was set to 0.0001, which was reduced by half every 20 epochs.

\subsection{Experimental Results}
\noindent\textbf{1) Result on GoPro.} We train our model on the training set of GoPro and evaluate it on the test set, as shown in Table \ref{table1}. Note that CSE metric is calculated based on the deblurred image and ground-truth for evaluating the performance of each deep unsupervised deblur model on each dataset, and we use the ratio of the CSE result of other methods to our model for comparison. We see that our FCL-GAN substantially outperforms the existing deep unsupervised BID methods in terms of model effectiveness, and meanwhile obtains a significant advantage in inference time and model size.

\noindent\textbf{2) Result on HIDE.} To quantitatively and qualitatively compare the generalization ability of each model, we directly apply the model pre-trained on GoPro to the HIDE dataset. As shown in Table \ref{table1} and Figure \ref{fig9}, our proposed FCL-GAN has more strong generalization ability and outperforms all data-driven deep unsupervised methods. We are also surprised that our FCL-GAN obtains highly competitive results from some deep supervised deblur models.

\noindent\textbf{3) Result on CelebA.} To evaluate the ability of each method for processing domain-specific blurred images, we train and test each model on CelebA face dataset. From Table \ref{table2} and Figure \ref{fig10}, our method achieves the optimal performance and visual effects among all unsupervised methods again, and deep unsupervised methods tend to outperform some deep supervised methods in this case.

\noindent\textbf{4) Result on RealBlur.} To examine the ability of each model to handle real-world blurs, we directly apply the model pre-trained on GoPro to the RealBlur dataset, as shown in Table \ref{table3} and Figure \ref{fig11}. From Table \ref{table3}, we see that our method outperforms all unsupervised methods. Figure \ref{fig11} shows that our method can better handle real-world blurring images.

\begin{table}[t]
	\centering
	\caption{Performance comparison on RealBlur \cite{rim2020real}.}
	\vspace{-3mm}
	\setlength{\tabcolsep}{0.5mm}{
		\begin{tabular}{l|l|cc|cc}
			\hline
			\multicolumn{2}{c|}{Datasets} & \multicolumn{2}{|c|}{RealBlur-J} & \multicolumn{2}{|c}{RealBlur-R} \\
			\hline
			\multicolumn{2}{c|}{Metrics} & PSNR/SSIM$\uparrow$ & CSE$\downarrow$  & PSNR/SSIM$\uparrow$ & CSE$\downarrow$ \\
			\hline
			Super- & DeblurGAN \cite{kupyn2018deblurgan} & 27.97/0.834 & / & 33.79/0.903 & / \\
			vised & DeepDeblur \cite{nah2017deep} & 27.87/0.827 & / & 32.51/0.841 & / \\
			\hline
			& CycleGAN \cite{zhu2017unpaired} & 19.79/0.633 & 2.898 & 12.38/0.242 & 3.024 \\
			Unsup- & UID-GAN \cite{lu2019unsupervised} & 22.87/0.671 & 1.394 & 16.64/0.323 & 2.985 \\
			ervised & Ours & \bfseries{25.35/0.736} & \bfseries{1} & \bfseries{28.37/0.663} & \bfseries{1} \\
			\hline
	\end{tabular}}
	\label{table3}
\end{table}

\begin{table}[t]
	\centering
	\caption{Ablation studies for loss functions on the GoPro dataset \cite{nah2017deep} and CelebA dataset \cite{liu2015deep}.}
	\vspace{-3mm}
	\setlength{\tabcolsep}{1.4mm}{
		\begin{tabular}{l|cccc}
			\hline
			Datasets & Model & $ w/o \mathcal{L}_{ctst}$ & $ w/o\mathcal{L}_{tv}$ & Ours \\
			\hline
			GoPro & PSNR/SSIM & 24.56/0.749 & 24.73/0.765 & \bfseries{24.84/0.771} \\
			\hline
			CelebA & PSNR/SSIM & 20.83/0.648 & 21.01/0.651 & \bfseries{21.07/0.652} \\
			\hline
	\end{tabular}}
	\label{table4}
\end{table}

\subsection{Ablaiton Studies}
\noindent\textbf{1) Effectiveness of loss function.} To verify the effectiveness of the used loss functions in training, we ablate $\mathcal{L}_{tv}$ and $\mathcal{L}_{ctst}$ on the GoPro and CelebA datasets. From the experimental results in Table \ref{table4}, we see that both $\mathcal{L}_{tv}$ and $\mathcal{L}_{ctst}$ have positive effects, and removing $\mathcal{L}_{ctst}$ has a more negative impact on the performance. 

\noindent\textbf{2) Effectiveness of designed basic meta and residual meta.} In this study, we design three different forms for verification, as shown in Figure \ref{fig4}. For the basic meta, we consider the following three forms: (a) Conv-BN-ReLU, (b) Conv-IN-ReLU, and (c) Conv-ReLU. While for the residual meta, we also consider three forms: (d) is the basic form of resblock \cite{he2016deep}; (e) is a simplified version of (d); (f) is the most widely used form in the field of deblurring \cite{nah2017deep,zhang2019deep,cho2021rethinking}. Table \ref{table5} shows the performance of various collaborative effects of basic meta and residual meta. We have the following conclusions which can be applied to unsupervised deblurring: 1) for the gain on performance: BN > norm-free > IN; 2) the introduction of IN substantially degrades the performance; 3) the simplified version of Residual meta performs better. Besides, norm-free seems to have the same effect as BN. However, norm-free leads to instability of training in experiments, as shown in Figure \ref{fig4} (g).

\begin{table}[t]
	\centering
	\caption{Ablation studies for different combinations of basic metas and residual metas on GoPro \cite{nah2017deep}.}
	\vspace{-3mm}
	\setlength{\tabcolsep}{0.6mm}{
		\begin{tabular}{l|cccc}
			\hline
			\bfseries{Combination} & (a)+(d) & (a)+(e) & (b)+(d) & (b)+(e) \\
			\bfseries{PSNR/SSIM} & 23.61/0.736 & \textbf{24.84/0.771} & 19.40/0.635 & 20.11/0.641\\
			\hline
			\bfseries{Combination} & (c)+(d) & (c)+(e) & (c)+(f) & \\
			\bfseries{PSNR/SSIM} & 23.47/0.729 & 24.55/0.765 & 24.20/0.762 &\\
			\hline
	\end{tabular}}
	\label{table5}
\end{table}

\begin{table}[t]
	\centering
	\caption{Ablation studies for structures on GoPro \cite{nah2017deep}.}
	\vspace{-3mm}
	\begin{tabular}{l|ccc}
		\hline
		\bfseries{Structures} & \bfseries{PSNR/SSIM} & \bfseries{Model Size} & \bfseries{Runtime} \\
		\hline
		\bfseries{\textit{Encoder-decoder}} & 24.66/0.763 & 50.30 (MB) & 15.19 (ms)\\
		\bfseries{\textit{Single-scale}} & 24.69/0.770 & 50.30 (MB) & 76.94 (ms)\\
		\bfseries{Our \textit{LED}} & \bfseries{24.84/0.771} & \bfseries{24.56} (MB) & \bfseries{10.86} (ms)\\
		\hline
	\end{tabular}
	\label{table6}
\end{table}

\noindent\textbf{3) Effectiveness of \textit{LED} structure.} We compare the deblurring performance, model size, and inference time of the three structures in Figure \ref{fig5}. Table \ref{table6} describes the comparison results. As can be seen, our LED is more light, infers faster, and performs better. 

\section{Conclusion}
We have discussed the limitations of existing unsupervised deep BID methods, and technically proposed a lightweight and real-time unsupervised BID baseline (FCL-GAN). We analyze the blurred and sharp images in the frequency domain and introduce frequency-domain contrastive learning to obtain superior performance. Qualitative and quantitative results show that our method achieves SOTA performance for the unsupervised BID and is even highly competitive with some supervised deep BID models on the domain-specific case. In terms of lightweight and real-time inference performance, our FCL-GAN method outperforms all existing image deblurring models (no matter supervised or unsupervised). Furthermore, our method only needs 0.011s to process a high-resolution image (1280x720) on an Nvidia RTX 2080Ti. In future, we will consider the deployment issue of our lightweight model, and also explore new stategies to furher imrpove the deblurring performance. 


\section{Acknowledgments}
This work is partially supported by the National Natural Science Foundation of China (62072151, 61732007, 61932009 and 62020106007), and the Anhui Provincial Natural Science Fund for Distinguished Young Scholars (2008085J30). Zhao Zhang is the corresponding author of this paper.

\bibliographystyle{ACM-Reference-Format}
\bibliography{FCL-GAN}

\end{document}